# IC-GVINS: A Robust, Real-time, INS-Centric GNSS-Visual-Inertial Navigation System for Wheeled Robot

Hailiang Tang, Tisheng Zhang, Xiaoji Niu, Jing Fan, and Jingnan Liu

***Abstract*—In this letter, we present a robust, real-time, inertial navigation system (INS)-Centric GNSS-Visual-Inertial navigation system (IC-GVINS) for wheeled robot, in which the precise INS is fully utilized in both the state estimation and visual process. To improve the system robustness, the INS information is employed during the whole keyframe-based visual process, with strict outlier-culling strategy. GNSS is adopted to perform an accurate and convenient initialization of the IC-GVINS, and is further employed to achieve absolute positioning in large-scale environments. The IMU, visual, and GNSS measurements are tightly fused within the framework of factor graph optimization. Dedicated experiments were conducted to evaluate the robustness and accuracy of the IC-GVINS on a wheeled robot. The IC-GVINS demonstrates superior robustness in various visual-degenerated scenes with moving objects. Compared to the state-of-the-art visual-inertial navigation systems, the proposed method yields improved robustness and accuracy in various environments. We open source our codes combined with the dataset on GitHub[1].**

## I. Introduction

Continuous, robust, and accurate positioning is an essential task for autonomous mobile robots, such as wheeled robots and aerial robots, in large-scale challenging environments [1]. Visual-inertial navigation system (VINS) has become an applicable solution for autonomous mobile robots, due to its higher accuracy and lower cost [2]. However, it has been historically difficult to achieve a robust and reliable positioning for VINS in complex environments, mainly because that the visual system is extremely sensitive to illumination change and moving objects [3]. In contrast, inertial measurement unit (IMU) would not be affected by these external environment factors, and inertial navigation system (INS) can maintain continuous high-frequency positioning independently [4]. A single low-cost micro-electro-mechanical system (MEMS) INS cannot provide long-term (e.g. longer than 1 minute) high-accuracy positioning, but it can actually achieve decimeter-level positioning within several seconds according to our experiments [5]. However, most of current VINSs are visual-centric or visual-driven, while the INS has not been well considered, such as in [6]-[7]. Furthermore, the INS make few or even no contributions to the visual process in these systems, which might degrade robustness and accuracy in visual-degenerated environments. In this letter, in order to fully utilize the advantages of the INS and finally improve the robustness and accuracy of the VINS, we propose an INS-centric visual-inertial navigation system (IC-VINS). We further incorporate the global navigation satellite system (GNSS) into the proposed IC-VINS to construct a GNSS-visual-inertial navigation system (GVINS), so as to achieve continuous, robust, and accurate absolute positioning in large-scale challenging environments.

Conventionally, the state-estimation in VINS is addressed through filtering, where the IMU measurements are propagated, and the visual measurements are adopted to update the system states [8]-[10]. The cloned INS pose are also used to triangulate the feature landmarks [8]-[9] in multi-state constraint Kalman filter (MSCKF). However, we have still noticed some insufficient usage of the INS in recent filtering-based approaches. Taking OpenVINS [8] for example, their implementation is visual-driven, because the system will pause if no image is received. However, the independent INS should be adopted for real-time navigation without hesitation. In addition, the INS make no contribution to the feature tracking in [8], which might degrade robustness in visual-degenerated environments. Similarly, an IEKF-based visual-inertial odometry (VIO) was proposed [11], in which the direct image intensity patches were employed as landmark descriptors allowing for tracking non-corner features. R-VIO [12] is a robocentric visual-inertial odometry within a MSCKF framework, which achieves competitive performance with state-of-the-art (SOTA) VINS. Though filtering-based VINSs have exhibited considerable accuracy in state estimation, they theoretically suffer from large linearization errors, which might possibly ruin the estimator and further degrade robustness and accuracy [13].

By solving maximum a posterior (MAP) estimation, factor graph optimization (FGO) has been proven to be more efficient and accurate than the filtering-based approaches [2], [13] for VINS. However, the INS has not been made full use in most of the FGO-based VINSs, and the IMU measurements have only

This research is funded by the National Key Research and Development Program of China (No. 2020YFB0505803), and the National Natural Science Foundation of China (No. 41974024). *(Corresponding authors: Xiaoji Niu; Tisheng Zhang.)*

Hailiang Tang, Tisheng Zhang, Xiaoji Niu, Jing Fan, and Jingnan Liu are with the GNSS Research Center, Wuhan University, Wuhan 430079, China (e-mail: {thl, zts, xjniu, jingfan, jnliu}@whu.edu.cn).

[1] https://github.com/i2Nav-WHU/IC-GVINS

been used to construct a relative constrain factor, such as IMU preintegration factor [5]-[7], [14]-[15]. VINS-Mono [6] adopts sliding-window optimization to achieve pose estimation, but their estimator relies more on the high-frequency visual observations. Though the latest pose integrated by IMU is output in real-time, the INS mechanization in [6] is imprecise, which is not suitable for high-accuracy positioning. In addition, their visual process [6] is relatively rough, which actually limits its accuracy in large-scale challenging environments. In [7], the camera pose predicted by INS is used to assist the ORB feature tracking instead of using the unreliable ad-hoc motion mode. The system in [7] is still driven by the visual image, and thus it is not suitable for real-time navigation. Similarly, Kimera-VIO [14] is a keyframe-based MAP visual-inertial estimator, which can perform both full smoothing or fixed-lag smoothing using GTAM [16], and their stereo VIO outperforms SOTA pipelines on public dataset, due to their delicate visual processes. A novel approach is proposed in [15], which combines the strengths of accurate VIO with globally consistent keyframe-based bundle adjustment (BA). Their works [15] is built upon the reality that the INS accuracy might quickly degrade after several seconds of integration. However, the INS can maintain decimeter-level positioning within several seconds [5], even for MEMS IMU, as mentioned above. As we can see, the INS is not well considered in these optimization-based VINSs, and the INS algorithm including the IMU-preintegration algorithm is rough. They actually waste the IMU precision to certain extend and finally degrade the accuracy of VINS. The high-accuracy industrial-grade MEMS IMU has been widely used for autonomous robot navigation, mainly because the cost of MEMS IMU has been lower and lower with improved accuracy [4]. In addition, most of these VINSs are driven by visual image, and are not suitable for real-time applications, which need stable and continuous positioning. Due to these reasons, we believe that the independent INS can play a more important role in both the state estimation and visual process of VINS, so as to improve the robustness and accuracy.

As we all know, GNSS can achieve absolute positioning in large-scale environments, and thus GNSS receiver is a common sensor for outdoor autonomous robots [4]. By using real-time kinematic (RTK) [4], [17], GNSS can even perform centimeter-level positioning in open-sky environments. In [18], the GNSS is integrated into a global estimator, while the local estimator is a VINS. The GNSS can help to estimate the IMU biases, but the GNSS is separated from the VINS estimator in [18]. The GNSS raw measurements are tightly incorporated into a VINS in [19], which can provide global estimation under indoor-outdoor environments. The approach in [19] is based on [6], but the visual processes have not been improved. Hence, [19] might also degrade robustness and accuracy in GNSS-denied environments. The GNSS can also help to initialize the VINS. In [20], the GNSS/INS integration and VINS are launched simultaneously to initialize a GNSS-visual-inertial navigation system for land vehicle, but the approach is loosely coupled. G-VIDO [21] is a similar system, but they further incorporate the vehicle dynamic to improve accuracy for autonomous driving. In [22], a tightly coupled optimization-based GNSS-Visual-Inertial odometry is proposed, but the GNSS make no contribution to the initialization of the visual system. The GNSS works in a different world frame from the VIO system in all these system [18]-[22], and the VIO has to be initialized separately. Nevertheless, the GNSS can help to initialize the INS first, and further to initialize the VINS, and finally they can work in a unified world frame without any transformation.

As can be seen, the INS can independently provide precise and high-frequency pose in short-term, and would not affected by external environment factors. Inspired by these advantages of the INS, we propose an INS-centric visual-inertial navigation system within an FGO framework, in which the INS is made full use in both the state estimation and visual process. The GNSS is also integrated into the proposed system, so as to achieve an accurate initialization and further perform absolute positioning. The main contributions of our work are as follows:

- We propose a real-time INS-centric visual-inertial navigation system for wheeled robot, which make full use of the precise INS information in both the state estimation and visual process. The IC-VINS is a keyframe-based estimator under the framework of FGO, with strict outlier-culling strategy in both the front-end and back-end.
- The GNSS is directly incorporated into the proposed IC-VINS to construct a GNSS-visual-inertial navigation system in a unified world frame, which can achieve an accurate and convenient initialization and perform absolute positioning in large-scale environments.
- Dedicated experiment results indicate that the proposed method can run in various visual-degenerated scenes with moving objects, and yields improved robustness and accuracy compared to the SOTA methods in various environments. The proposed method also exhibits superior real-time performance.
- We open source our implementation on GitHub, together with our well-synchronized multi-sensor dataset collecting by a wheeled robot.

## II. System Overview

The proposed IC-GVINS for wheeled robot is depicted in Fig. 1. The whole system is driven by a precise INS mechanization. A GNSS/INS integration is conducted first to initialize the INS, so as to get rough IMU biases and absolute initial attitude estimation. The absolute attitude is aligned to the local navigation frame (gravity aligned) [4]-[5]. After the INS is initialized, the prior pose derived by the INS are directly employed for feature tracking, keyframe selection, and landmark triangulation. The IMU, visual, and GNSS measurements are tightly fused within an FGO framework to achieve MAP estimation. The estimated states are fed back to the INS mechanization module to update the newest INS states for real-time navigation. The wheeled odometer can be also incorporated into the FGO to further improve the robustness and accuracy. With the INS-centric scheme, the proposed IC-GVINS can provide continuous, robust, accurate positioning in large-scale complex environments for wheeled robot.

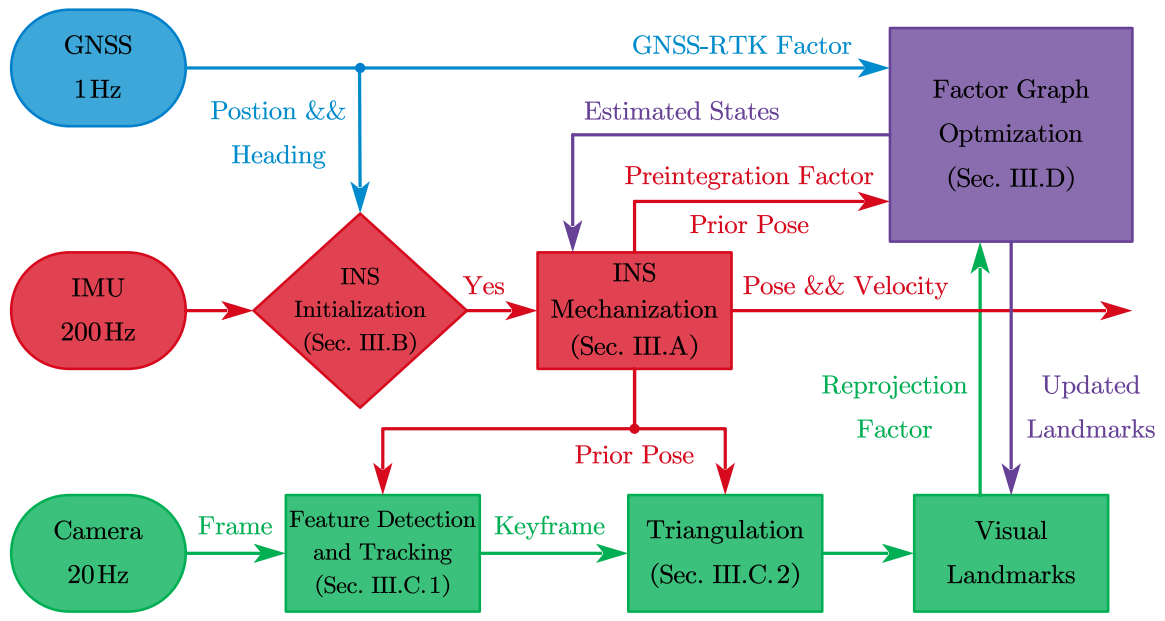

Fig. 1. System overview of the IC-GVINS.

## III. Methodology

In this section, the methodology of the proposed IC-GVINS is presented. The system core is a precise INS mechanization with the Earth rotation considered. A GNSS/INS integration is conducted first, so as to initialize the INS. Then, the visual process is aided by the prior pose derived from INS. Finally, all the measurements are fused together using FGO to achieve MAP estimation.

### A. INS Mechanization

The Earth rotation is not a negligible factor for INS, especially for industrial-grade or higher-grade MEMS IMUs [5]. To fully utilize the INS precision, we follow our previous work in [5] to adopt a precise INS mechanization algorithm, which consider the Earth rotation and the Coriolis acceleration [4]. The INS kinematic model is defined as follows:

$$\begin{aligned}
&\dot{\boldsymbol{p}}_{\mathrm{wb}}^{\mathrm{w}}=\boldsymbol{v}_{\mathrm{wb}}^{\mathrm{w}},\\
&\dot{\boldsymbol{v}}_{\mathrm{wb}}^{\mathrm{w}}=\mathbf{R}_{\mathrm{b}}^{\mathrm{w}}\boldsymbol{f}^{\mathrm{b}}+\boldsymbol{g}^{\mathrm{w}}-2[\boldsymbol{w}_{\mathrm{ie}}^{\mathrm{w}}\times]\boldsymbol{v}_{\mathrm{wb}}^{\mathrm{w}},\\
&\dot{\mathbf{q}}_{\mathrm{b}}^{\mathrm{w}}=\frac{1}{2}\mathbf{q}_{\mathrm{b}}^{\mathrm{w}}\otimes\begin{bmatrix}0\\ \boldsymbol{w}_{\mathrm{wb}}^{\mathrm{b}}\end{bmatrix},\ \boldsymbol{w}_{\mathrm{wb}}^{\mathrm{b}}=\boldsymbol{w}_{\mathrm{ib}}^{\mathrm{b}}-\mathbf{R}_{\mathrm{w}}^{\mathrm{b}}\boldsymbol{w}_{\mathrm{ie}}^{\mathrm{w}},
\end{aligned} \tag{1}$$

where the world frame (w-frame) is defined at the initial position of the navigation frame (n-frame) or the local geodetic north-east-down (NED) frame; the IMU frame is defined as the body frame (b-frame); $\boldsymbol{g}^{\mathrm{w}}$ and $\boldsymbol{w}_{\mathrm{ie}}^{\mathrm{w}}$ are the gravity vector and the Earth rotation rate in the w-frame; $\mathbf{R}_{\mathrm{b}}^{\mathrm{w}}$ is the rotation matrix corresponding to the quaternion $\mathbf{q}_{\mathrm{b}}^{\mathrm{w}}$. The precise INS mechanization can be formulated by adopting the kinematic model in (1). For more details about the precise INS mechanization, one can refer to [5]. The integrated pose by INS mechanization is directly used for real-time navigation, and also provide prior aiding for the visual process, as depicted in Fig. 1.

### B. Initialization

The initialization is an essential procedure for VINS, which determines the system's robustness and accuracy [6]-[7]. As an INS-centric system, the most important task is to initialize the INS. Inspired by the GNSS/INS initialization, a GNSS/INS integration within the FGO framework is adopted to initialize the INS, and the FGO is described in section III.C. After the INS initialization, we obtain a rough IMU biases and absolute attitude estimation. The absolute attitude is essential for the IC-GVINS, mainly because we can incorporate the GNSS directly without heading alignment or coordinate transformation, and the precise IMU preintegration also needs absolute attitude to consider the Earth rotation [5]. By detecting the zero-velocity states, we can also obtain a rough estimation of roll, pitch, and gyroscope biases during stationary state [23].

The initialized INS can provide prior pose for visual process, which directly initialize the visual system. Once the landmarks have been triangulated, the visual reprojection factors can be constructed using the visual observations. A joint optimization is conducted to further refine the state estimation, and improve the following INS precision. According to our experiments, only 5 seconds' GNSS (in dynamic condition) is needed to perform an accurate initialization for the proposed method, while the time length is 9 seconds in [20]. Once the initialization has been finished, the INS-centric VINS can work independently without GNSS.

### C. INS-Aided Visual Process

The proposed IC-VINS is a keyframe-based visual-inertial navigation system. The prior pose from the INS are fully used in the entire visual process, including the feature tracking and triangulation, so as to improve the system robustness and accuracy. Strict outlier-culling strategy is conducted to further improve the robustness and accuracy.

#### 1) Feature Detection and Tracking

Shi-Tomasi corner features are detected in our visual front-end. The image is first divided into several grids with a setting size, e.g. 200 pixels. The features are detected separately in each grid, and a minimum separation of two neighboring pixels is also set, so as to maintain a uniform distribution of the features. Multi-thread technology is employed to improve the detection efficiency. Lukas-Kanade optical flow algorithm is adopted to track the features. For those features without initial depth, they are tracked directly, and RANSAC is employed to reject outliers. For those features with depth, initial estimations are first predicted using the prior pose from the INS, and then they are tracked with the initial optical flows. We also track the features in backward direction (from current frame to previous frame), and further remove the failed matches. The tracked features will be undistorted for further processes.

Once the features have been tracked, the keyframe selection is conducted. We first calculate the average parallax between the current frame and last keyframe. The prior pose from the INS is adopted to compensate the rotation during calculation, rather than the raw gyroscope measurements as in [6]. If the average parallax is larger than a fixed threshold, e.g. 20 pixels, then the current frame is selected as a new keyframe. The selected keyframe will be used to triangulate landmarks, and further to construct reprojection factors in FGO. However, if the robot states stationary or the average parallax is smaller than the threshold for a long time, no new optimization will be conducted in FGO, which might degrade the system accuracy. Hence, if no new keyframe is selected after a long time, e.g. 0.5 seconds, a new observation frame will be inserted into the keyframe queue. The observation frame will be used only for one time, and will be removed after the optimization.

#### 2) Triangulation

With the prior pose from the INS, triangulation has become a part of the visual front-end, which has facilitated the visual process. When a new keyframe is selected, triangulation will be conducted between the current keyframe and the previous keyframes. The triangulation determines the initial depth of the landmark, and the depth will be further estimated in the FGO. Hence, strict outlier-culling strategy is also conducted in triangulation, so as to prevent the outlier landmarks or poorly initialized landmarks to ruin the FGO estimator. Parallax is first calculated between the feature in current keyframe and corresponding feature in first observed keyframe. If the parallax is too small, e.g. 10 pixels, the feature will be tracked until the parallax is enough, which can improve the precision of the triangulated depths. Then, the prior pose from the INS is used to triangulate the landmarks, and the depth of the landmark in its first observed keyframe can be obtained. We further check the depths to ensure the correctness of the triangulation. Only those depths within a range, e.g. 1~100 meters, will be added to the landmark queue, or will be treated as outliers.

### D. *Factor Graph Optimization*

A sliding-window optimizer is adopted to tightly fuse the measurements under the framework of FGO. When a new keyframe is selected or a new GNSS-RTK measurement is valid, a new time node will be inserted into the sliding window, and factor graph optimization will be carried out to perform MAP estimation. It should be noted that time node is always inserted at GNSS seconds, whether the GNSS-RTK is valid at this time node or not. The IMU preintegration factor is constructed between each consecutive time node. The FGO framework of the proposed IC-GVINS is depicted in Fig. 2.

#### 1) *Formulation*

The state vector $\boldsymbol{X}$ in the sliding window of the IC-GVINS can be defined as

$$\begin{aligned} \boldsymbol{X} &= [\boldsymbol{x}_0,\ \boldsymbol{x}_1,\ \dots,\ \boldsymbol{x}_n,\ \boldsymbol{x}_c^b,\ \delta_0,\ \delta_1,\ \dots,\ \delta_l], \\ \boldsymbol{x}_k &= [\boldsymbol{p}^{\mathrm{w}}_{\mathrm{wb}_k},\ \mathbf{q}^{\mathrm{w}}_{\mathrm{b}_k},\ \boldsymbol{v}^{\mathrm{w}}_{\mathrm{wb}_k},\ \boldsymbol{b}_{g_k},\ \boldsymbol{b}_{a_k}],\ k\in[0,\ n], \\ \boldsymbol{x}_{\mathrm{c}}^{\mathrm{b}} &= [\boldsymbol{p}^{\mathrm{b}}_{\mathrm{bc}},\ \mathbf{q}^{\mathrm{b}}_{\mathrm{c}}], \end{aligned} \tag{2}$$

where $\boldsymbol{x}_k$ is the IMU state at each time node, as showed in Fig. 2; the IMU state includes position, attitude quaternion, and velocity in the w-frame, and gyroscope biases $\boldsymbol{b}_g$ and accelerometer biases $\boldsymbol{b}_a$; $n$ is the number of time node in the sliding window; $\boldsymbol{x}_{\mathrm{c}}^{\mathrm{b}}$ is the extrinsic parameters between the camera frame (c-frame) and the IMU b-frame; $\delta$ is the inverse depth parameter of the landmark in its first observed keyframe.

The MAP estimation in IC-GVINS can be formulated by minimizing the sum of the prior and the Mahalanobis norm of all measurements as

$$\min_{\boldsymbol{X}} \left\{ \begin{aligned} &\|\mathbf{r}_p - \mathbf{H}_p\boldsymbol{X}\|^2 + \sum_{k\in[1,\ n]} \|\mathbf{r}_{Pre}(\tilde{\boldsymbol{z}}^{Pre}_{k-1,\ k},\ \boldsymbol{X})\|^2_{\boldsymbol{\Sigma}^{Pre}_{k-1,\ k}} \\ &+ \sum_{l\in L} \|\mathbf{r}_V(\tilde{\boldsymbol{z}}^{V_{i,j}}_l,\ \boldsymbol{X})\|^2_{\boldsymbol{\Sigma}^{V_{i,j}}_l} \\ &+ \sum_{h\in[0,\ m]} \|\mathbf{r}_{GNSS}(\tilde{\boldsymbol{z}}^{GNSS}_h,\ \boldsymbol{X})\|^2_{\boldsymbol{\Sigma}^{GNSS}_h} \end{aligned} \right\}, \tag{3}$$

where $\mathbf{r}_{Pre}$ are the residuals of the IMU preintegration measurements; $\mathbf{r}_V$ are the residuals of the visual measurements; $\mathbf{r}_{GNSS}$ are the residuals of the GNSS-RTK measurements; $\{\mathbf{r}_p,\ \mathbf{H}_p\}$ represents the prior from marginalization [6]; $m$ is the number of GNSS-RTK measurements in the sliding window; $L$ is the landmark map in the sliding window, and $l$ is the landmark in the map; $i$ denotes the reference keyframe of the landmark $l$, and $j$ is another keyframe. The Ceres solver [25] is adopted to solve this FGO problem.

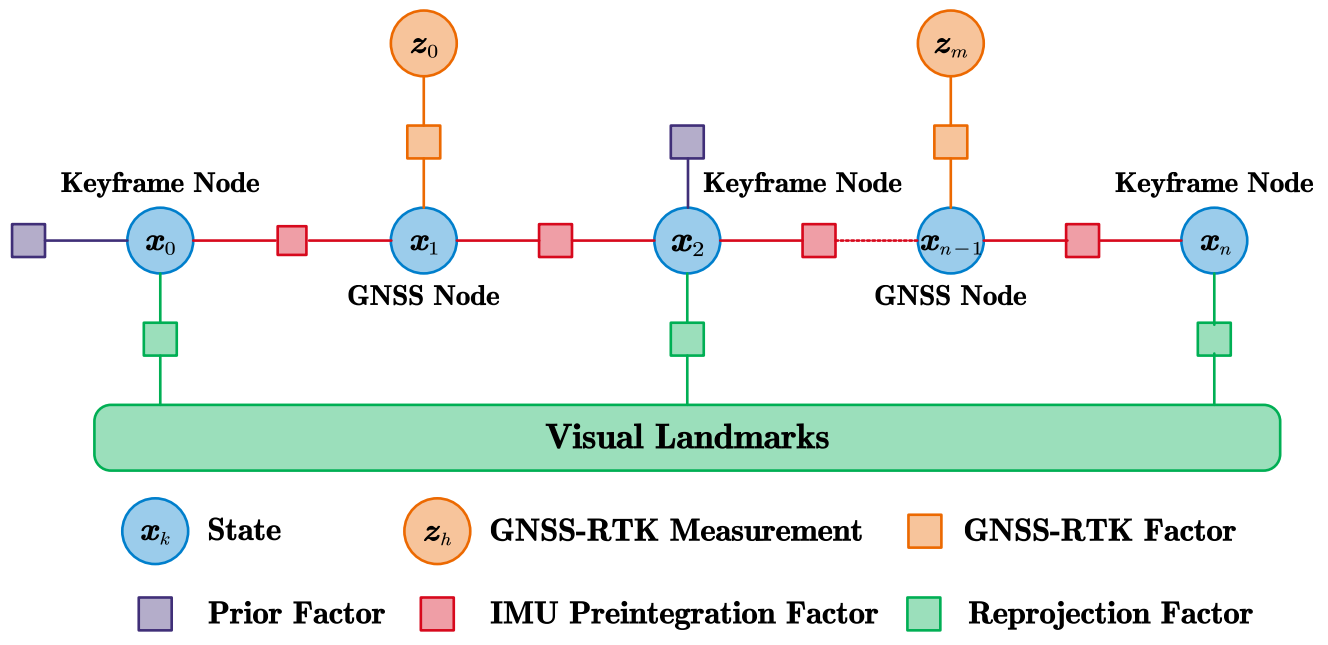

Fig. 2. FGO framework of the IC-GVINS.

#### 2) *IMU Preintegration Factor*

Compensating the Earth rotation has been proven to improve the accuracy of the IMU preintegration [5], and thus we follow our refined IMU preintegration in this letter. The residual of the IMU preintegration measurement can be written as

$$\mathbf{r}_{Pre}(\tilde{\boldsymbol{z}}^{Pre}_{k-1,k},\ \boldsymbol{X}) = \begin{bmatrix} (\mathbf{R}^{\mathrm{w}}_{\mathrm{b}_{k-1}})^T \begin{pmatrix} \boldsymbol{p}^{\mathrm{w}}_{\mathrm{wb}_k} - \boldsymbol{p}^{\mathrm{w}}_{\mathrm{wb}_{k-1}} - \boldsymbol{v}^{\mathrm{w}}_{\mathrm{wb}_{k-1}}\Delta t_{k-1,k} \\ -0.5\boldsymbol{g}^{\mathrm{w}}\Delta t^2_{k-1,k} + \Delta\boldsymbol{p}^{\mathrm{w}}_{g/cor,k-1,k} \end{pmatrix} - \Delta\hat{\boldsymbol{p}}^{Pre}_{k-1,k} \\ (\mathbf{R}^{\mathrm{w}}_{\mathrm{b}_{k-1}})^T \begin{pmatrix} \boldsymbol{v}^{\mathrm{w}}_{\mathrm{wb}_k} - \boldsymbol{v}^{\mathrm{w}}_{\mathrm{wb}_{k-1}} - \\ \boldsymbol{g}^{\mathrm{w}}\Delta t_{k-1,k} + \Delta\boldsymbol{v}^{\mathrm{w}}_{g/cor,k-1,k} \end{pmatrix} - \Delta\hat{\boldsymbol{v}}^{Pre}_{k-1,k} \\ 2\left[(\mathbf{q}^{\mathrm{w}}_{\mathrm{b}_k})^{-1} \otimes \mathbf{q}^{\mathrm{w}}_{\mathrm{w}_{\mathrm{i}(k-1)}}(t_k) \otimes \mathbf{q}^{\mathrm{w}}_{\mathrm{b}_{k-1}} \otimes \hat{\mathbf{q}}^{Pre}_{k-1,k}\right]_{\boldsymbol{v}} \\ \boldsymbol{b}_{g_k} - \boldsymbol{b}_{g_{k-1}} \\ \boldsymbol{b}_{a_k} - \boldsymbol{b}_{a_{k-1}} \end{bmatrix}, \tag{4}$$

where $\Delta\boldsymbol{p}^{\mathrm{w}}_{g/cor,k-1,k}$ and $\Delta\boldsymbol{v}^{\mathrm{w}}_{g/cor,k-1,k}$ are the Coriolis correction term for position and velocity preintegration [5], respectively; quaternion $\mathbf{q}^{\mathrm{w}}_{\mathrm{w}_{\mathrm{i}(k-1)}}(t_k)$ is the rotation caused by the Earth rotation. For more details about the refined IMU preintegration, one can refer to [5]. The wheeled odometer also can be integrated into the preintegration to further improve the system accuracy [24], which is also included in our source code.

#### 3) *Visual Reprojection Factor*

We follow [6], [19] to construct the visual reprojection factor in unit camera frame. The observed feature in pixel plane can be expressed as

$$\tilde{\boldsymbol{p}}_{\mathrm{p}} = [u,\ v]^T. \tag{5}$$

For a landmark $l$ with its inverse depth $\delta_l$ in first observed keyframe $i$, and another observed keyframe $j$, we can write the visual reprojection residual as

$$\begin{aligned} \mathbf{r}_V(\tilde{\boldsymbol{z}}^{V_{i,j}}_l,\ \boldsymbol{X}) &= \begin{bmatrix} \boldsymbol{b}_1 \\ \boldsymbol{b}_2 \end{bmatrix} \left( \frac{\hat{\boldsymbol{p}}_{\mathrm{c}_j}}{\|\hat{\boldsymbol{p}}_{\mathrm{c}_j}\|} - \pi_{\mathrm{c}}^{-1}(\tilde{\boldsymbol{p}}_{\mathrm{p}_j}) \right), \\ \hat{\boldsymbol{p}}_{\mathrm{c}_j} &= \mathbf{R}^{\mathrm{c}}_{\mathrm{b}} \left( \mathbf{R}^{\mathrm{b}_j}_{\mathrm{w}} \left( \begin{aligned} &\mathbf{R}^{\mathrm{w}}_{\mathrm{b}_i} \left( \mathbf{R}^{\mathrm{b}}_{\mathrm{c}} \frac{1}{\delta_l} \pi_{\mathrm{c}}^{-1}(\tilde{\boldsymbol{p}}_{\mathrm{p}_i}) + \boldsymbol{p}^{\mathrm{b}}_{\mathrm{bc}} \right) \\ &+ \boldsymbol{p}^{\mathrm{w}}_{\mathrm{wb}_i} - \boldsymbol{p}^{\mathrm{w}}_{\mathrm{wb}_j} \end{aligned} \right) - \boldsymbol{p}^{\mathrm{b}}_{\mathrm{bc}} \right), \end{aligned} \tag{6}$$

where $\pi_{\mathrm{c}}^{-1}$ is the back camera projection function, which transforms a feature in pixel plane $\boldsymbol{p}_{\mathrm{p}}$ into unit camera frame using the camera intrinsic parameters; $\mathbf{R}_{\mathrm{c}}^{\mathrm{b}}$ and $\boldsymbol{p}_{\mathrm{bc}}^{\mathrm{b}}$ are the extrinsic parameters between c-frame and b-frame, as in (2); $\mathbf{R}_{\mathrm{b}}^{\mathrm{w}}$ and $\boldsymbol{p}_{\mathrm{wb}}^{\mathrm{w}}$ are the pose parameters of the IMU expressed in the w-frame; $\boldsymbol{b}_1$ and $\boldsymbol{b}_2$ are two orthogonal bases that span the tangent plane of $\hat{\boldsymbol{p}}_{\mathrm{c}_j}$.

#### 4) *GNSS-RTK Factor*

The GNSS-RTK positioning in geodetic coordinates can be converted to the local w-frame as $\hat{\boldsymbol{p}}_{GNSS}^{\mathrm{w}}$ [4]. By considering the GNSS lever-arms $\boldsymbol{l}_{GNSS}^{\mathrm{b}}$ in b-frame, the residual of the GNSS-RTK measurement can be written as

$$\mathbf{r}_{GNSS}(\tilde{\boldsymbol{z}}_h^{GNSS}, \boldsymbol{X}) = \boldsymbol{p}_{\mathrm{wb}_h}^{\mathrm{w}} + \mathbf{R}_{\mathrm{b}_h}^{\mathrm{w}} \boldsymbol{l}_{GNSS}^{\mathrm{b}} - \hat{\boldsymbol{p}}_{GNSS,h}^{\mathrm{w}}. \quad (7)$$

As can be seen, the GNSS-RTK is directly incorporated into the FGO without extra coordinate transformation or yaw alignment as in [19]-[22], which is benefit from the INC-centric architecture.

#### 5) *Outlier Culling*

A two-step optimization is employed in the proposed IC-GVINS. After the first optimization, a chi-square test (degree of confidence of 95%) is conducted to remove all the unsatisfied visual reprojection factors from the optimizer, rather than the landmark map. The second optimization is then carried out to achieve a better state estimation. Once these two optimizations have been finished, the outlier-culling process is implemented. The position of the landmarks in w-frame are first calculated. The depth and reprojection error of each landmark are then evaluated in its observed keyframes. The unsatisfied feature observations, e.g. the depth is not within 1~100 meters or the reprojection error exceeds 4.5 pixels, will be marked as outliers, and will not be used in the next optimization. Furthermore, the average reprojection error of each landmark is calculated, and the landmark will be removed from the landmark map if the value is larger than the threshold, e.g. 1.5 pixels. As we can see, we not only remove landmark outliers, but also remove feature observation outliers, which significantly improve the system robustness and accuracy.

## IV. Experiments and Results

### A. *Implementation and Equipment setup*

The proposed IC-GVINS is implemented using C++ under the framework of Robot Operating System (ROS), which is suitable for real-time application. The dataset collected by a wheeled robot is adopted for the evaluation. The equipment setup of the wheeled robot is showed in Fig. 3. The sensors include a global shutter camera with the resolution of 1280x1024 (Allied Vision Mako-G131), an industrial-grade MEMS IMU (ADI ADIS16465), and a dual-antenna GNSS-RTK receiver (NovAtel OEM-718D). All the sensors have been synchronized through hardware trigger to the GNSS time. The intrinsic and extrinsic parameters of the camera have been well calibrated using the Kalibr [26] in advance. An on-board computer (NVIDIA Xavier) is employed to record the multi-sensor dataset. A navigation-grade [4] GNSS/INS integrated navigation system is adopted as the ground-truth system. The average speed of the wheeled robot is about 1.5 m/s.

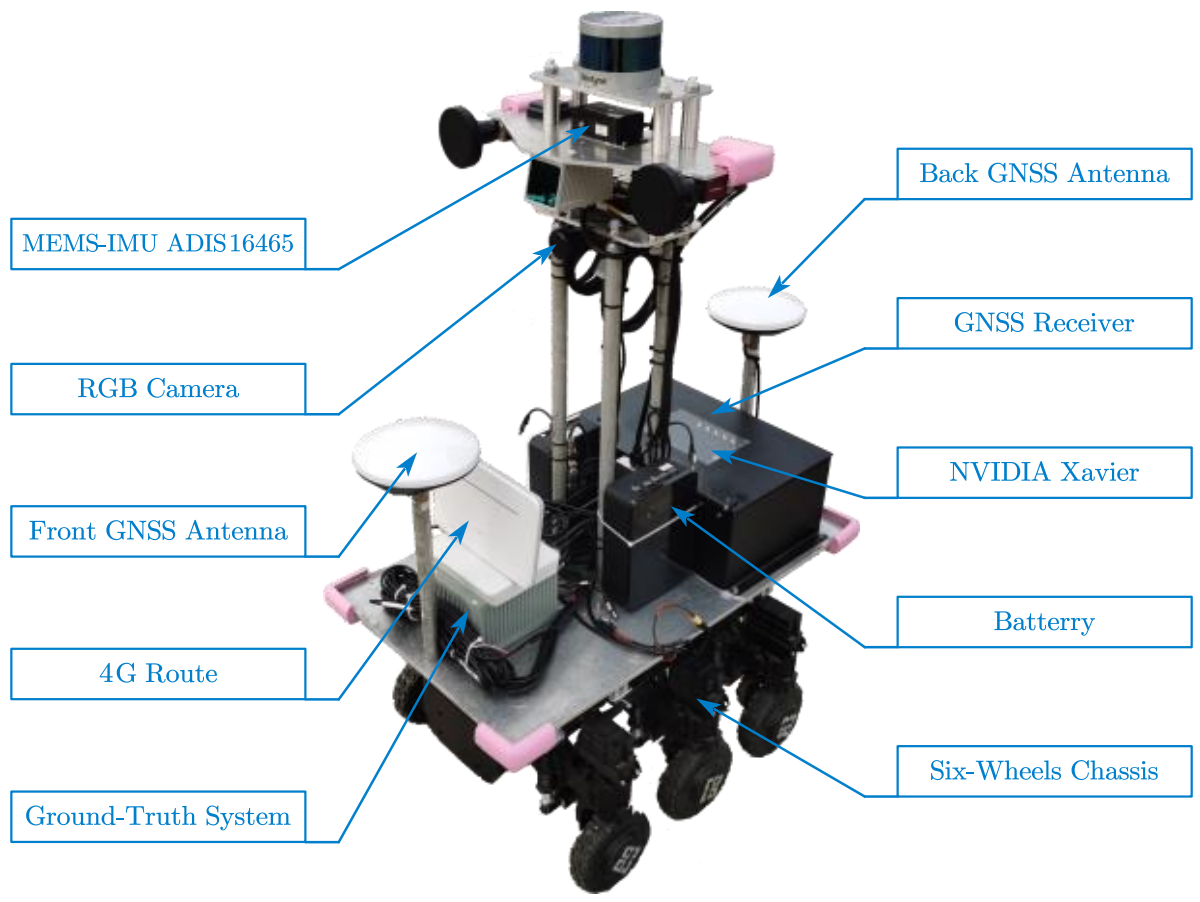

Fig. 3. Equipment setup of the wheeled robot.

We performed comparison with SOTA visual-inertial navigation systems VINS-Mono (without relocalization) [6], and OpenVINS [8]. The number of the maximum features for all the systems is set to 150 for justice. The temporal and spatial parameters between the camera and IMU are all estimated and calibrated online. We also conducted parameters tuning for VINS-Mono and OpenVINS on our dataset to achieve a better accuracy. Three experiments were conducted to fully evaluate the robustness and accuracy of the proposed method. The experiment-1 was carried out in various visual-degenerated environments with lots of moving objects to verify the system robustness. The experiment-2 was conducted to evaluate the accuracy of these VINS in an open-sky environment. The experiment-3 is the most significant experiment to demonstrate the robustness and accuracy of the proposed method, and two tests were carried out in large-scale challenging environments.

It should be noted that the proposed IC-VINS uses 5 seconds' GNSS-RTK for system initialization, as mentioned in section III.B, while the IC-GVINS uses all valid GNSS-RTK. Absolute error and relative error are adopted to quantitatively evaluated the accuracy [27]. For relative error, we calculated the relative rotation error (RRE) and the relative translation error (RTE) over some sub-sequences of length (50, 100, 150, and 200 meters). Evo [28] is adopted to quantitatively calculate the absolute and relative error. All the systems are running in real time under the framework of ROS.

### B. *Experiment-1: Robustness evaluation in visual-degenerated environments with moving objects*

In this experiment, the robustness of the proposed system was qualitatively evaluated in visual-degenerated environments with lots of moving objects. This experiment was conducted around Xinghu building group in Wuhan University, where GNSS-RTK tended to have few fixed solutions, and thus there was no valid ground truth in this experiment, as depicted in Fig. 4. The trajectory length in this experiment is 1151 meters and 832 seconds. There are lots of challenging scenes in this experiment: 1) drastic illumination change when crossing the tunnel, which might result in less tracked features; 2) repetitive textures from the building, which might cause false tracking; 3) lots of moving objects, including pedestrians, bicycles, and vehicles, which might cause landmark outliers and further ruin the optimizer. As showed in Fig. 5, the proposed method exhibits superior robustness in such challenging scenes, while VINS-Mono and OpenVINS have large drifts. The differences are especially significant at the narrow corridors, as depicted in Fig. 5. This experiment indicates that the INS-centric scheme together with the strict outlier-culling strategy can significantly improve the robustness of the proposed method in visual-degenerated environments.

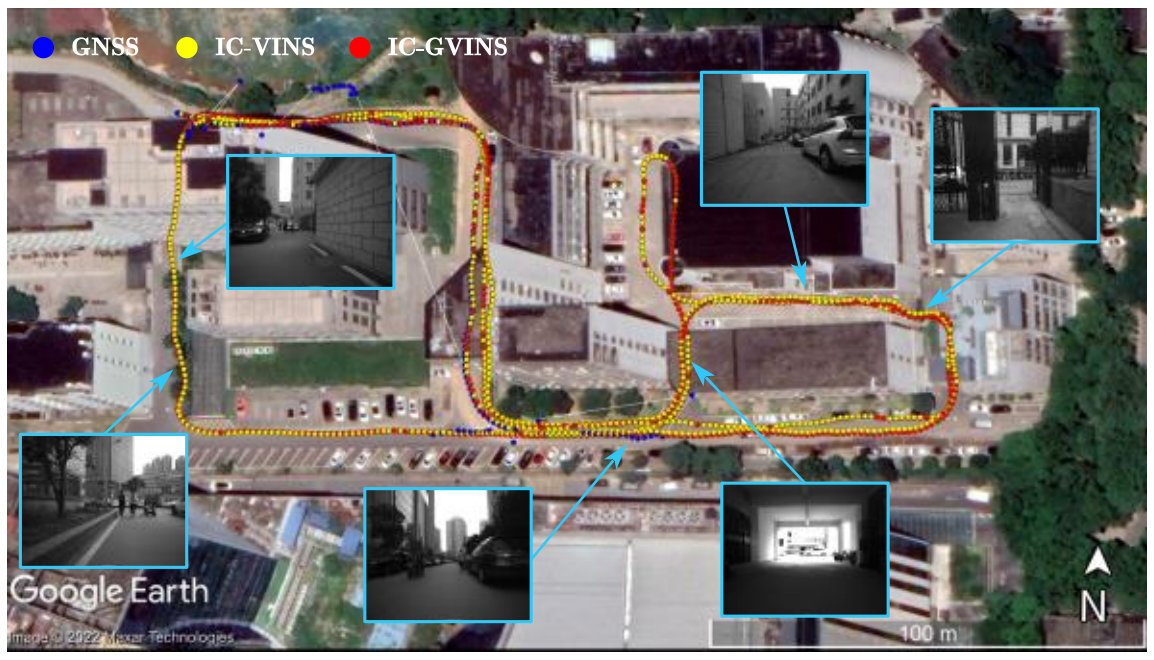

Fig. 4. The degenerated scenes in experiment-1.

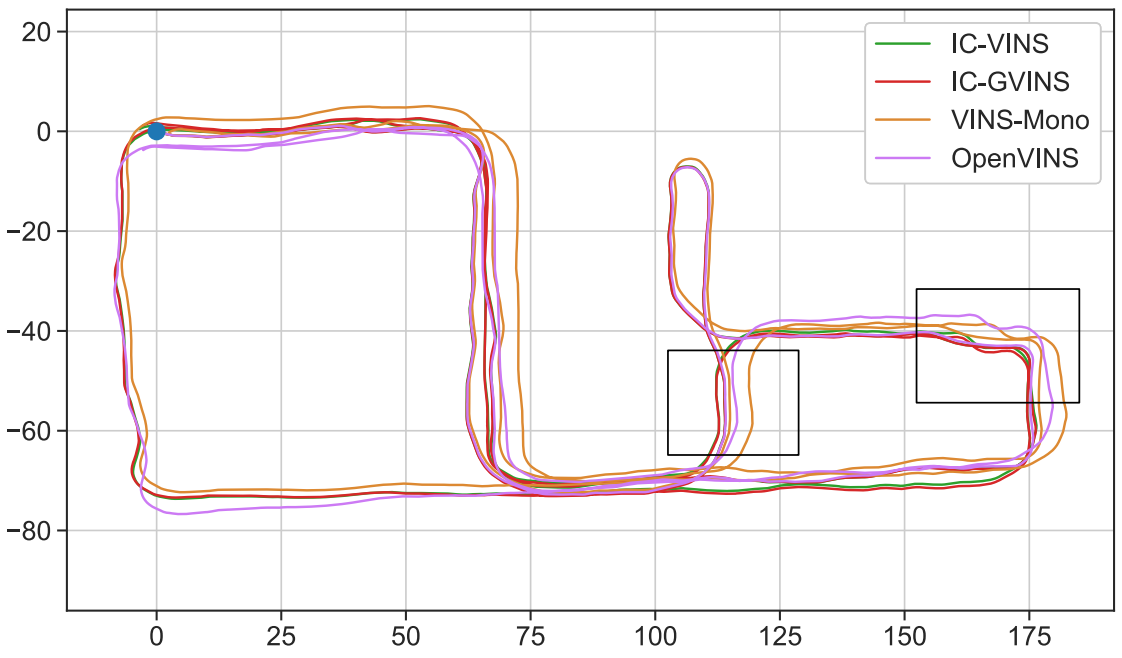

Fig. 5. The test trajectories in experiment-1. The black rectangles in the figure correspond to the two narrow corridors. The blue circle represents the start point. The trajectory length is 1151 meters.

### C. *Experiment-2: Quantitative evaluation in an open-sky environment*

This experiment was conducted in an abandoned playground with a huge mound in the middle, where GNSS-RTK can achieve centimeter-level positioning all the time, as depicted in Fig. 6. Rich visual textures are around the test trajectory, and there are nearly no moving objects in this test scene. The trajectory length in this experiment is 1657 meters and 1169 seconds. As can be seen in Fig. 6 and Fig. 7, the trajectories of the proposed method are well aligned to the ground truth, though the IC-VINS appears small drifts. However, the VINS-Mono and OpenVINS have large drifts, even though their parameters have been tuned. The RRE and RTE in this experiment are showed in Table I. The proposed IC-VINS outperforms VINS-Mono and OpenVINS in both RRE and RTE, especially for the rotation accuracy and for the long-term accuracy, which benefits from the INS-centric architecture and the precise INS mechanization.

### D. *Experiment-3: Quantitative evaluation in large-scale challenging environments*

In this experiment, two tests were conducted in large-scale challenging environments, i.e. complex campus scenes, to further evaluate the robustness and accuracy of the proposed method. The test scenes are surrounded by quantities of trees and buildings, as depicted in Fig. 8. There are also lots of fast-moving objects, and even traffic on the avenue, which make this experiment extremely challenging. The trajectory length of the two tests, named experiment-3a and experiment-3b, are 1535 meters (1087 seconds) and 2554 meters (1801 seconds), respectively. As showed in Fig. 9 and Fig. 10, the trajectories of the proposed method are well aligned to the ground truth, while VINS-Mono and OpenVINS gradually deviate.

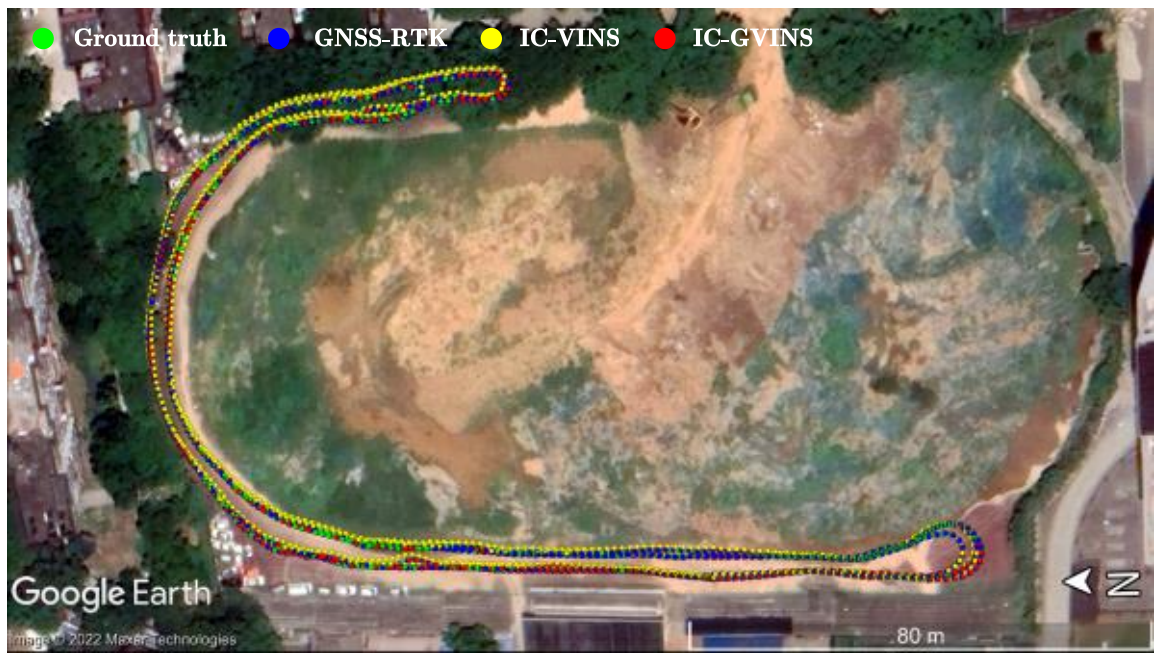

Fig. 6. The test scenes in experiment-2.

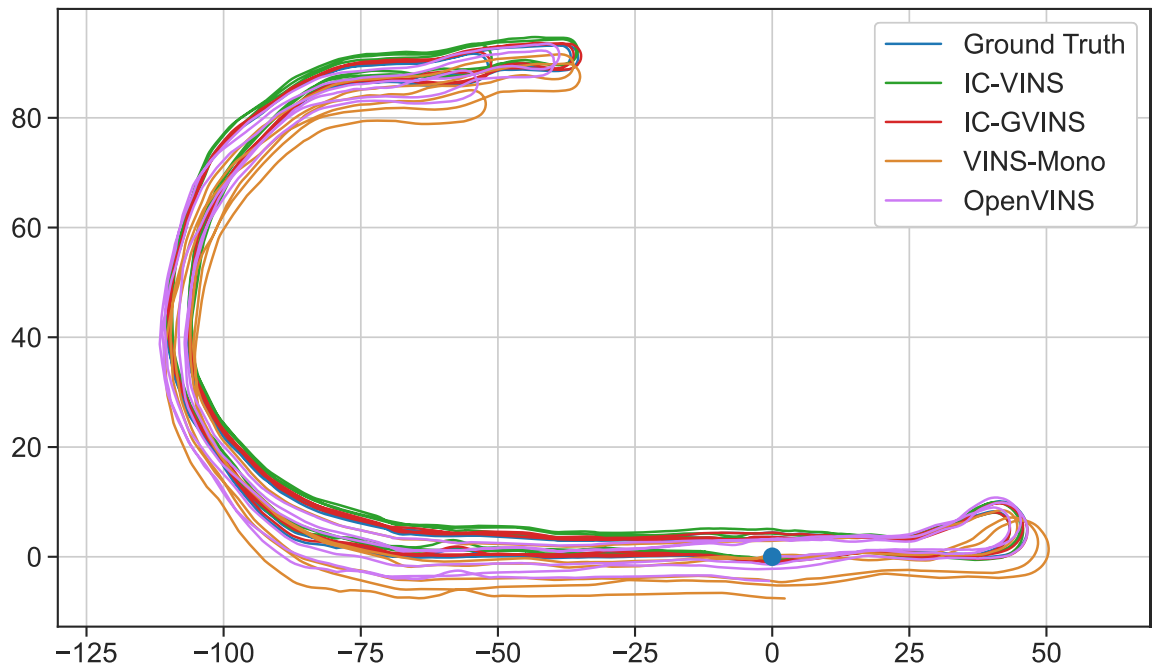

Fig. 7. The trajectories in experiment-2. The blue circle represents the start point. The trajectory length is 1657 meters.

TABLE I
RELATIVE ROTATION AND TRANSLATION ERROR IN EXPERIMENT-2

| RRE / RTE | 50m (deg / %) | 100m (deg / %) | 150m (deg / %) | 200m (deg / %) |
|---|---|---|---|---|
| VINS-Mono | 0.31 / 1.65 | 0.47 / 1.31 | 0.59 / 1.16 | 0.69 / 1.01 |
| OpenVINS | 0.43 / 0.83 | 0.59 / 0.71 | 0.72 / 0.70 | 0.76 / 0.67 |
| IC-VINS | **0.23 / 0.61** | **0.35 / 0.45** | **0.45 / 0.39** | **0.51 / 0.37** |

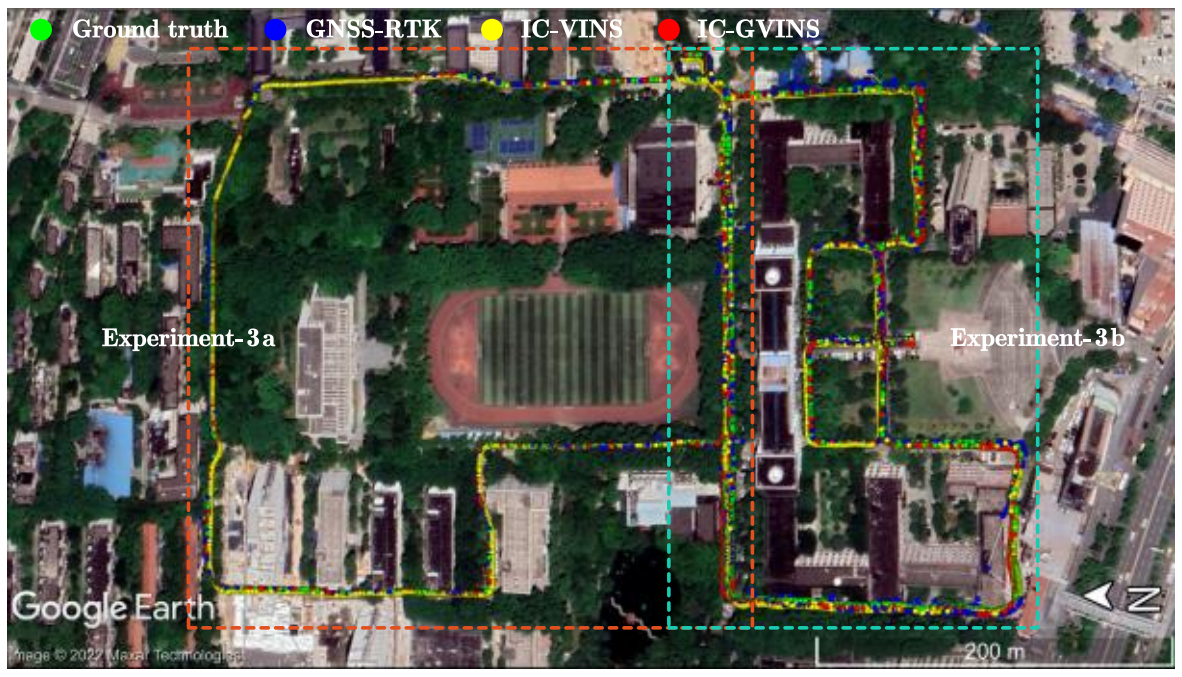

Fig. 8. The two test scenes in experiment-3.

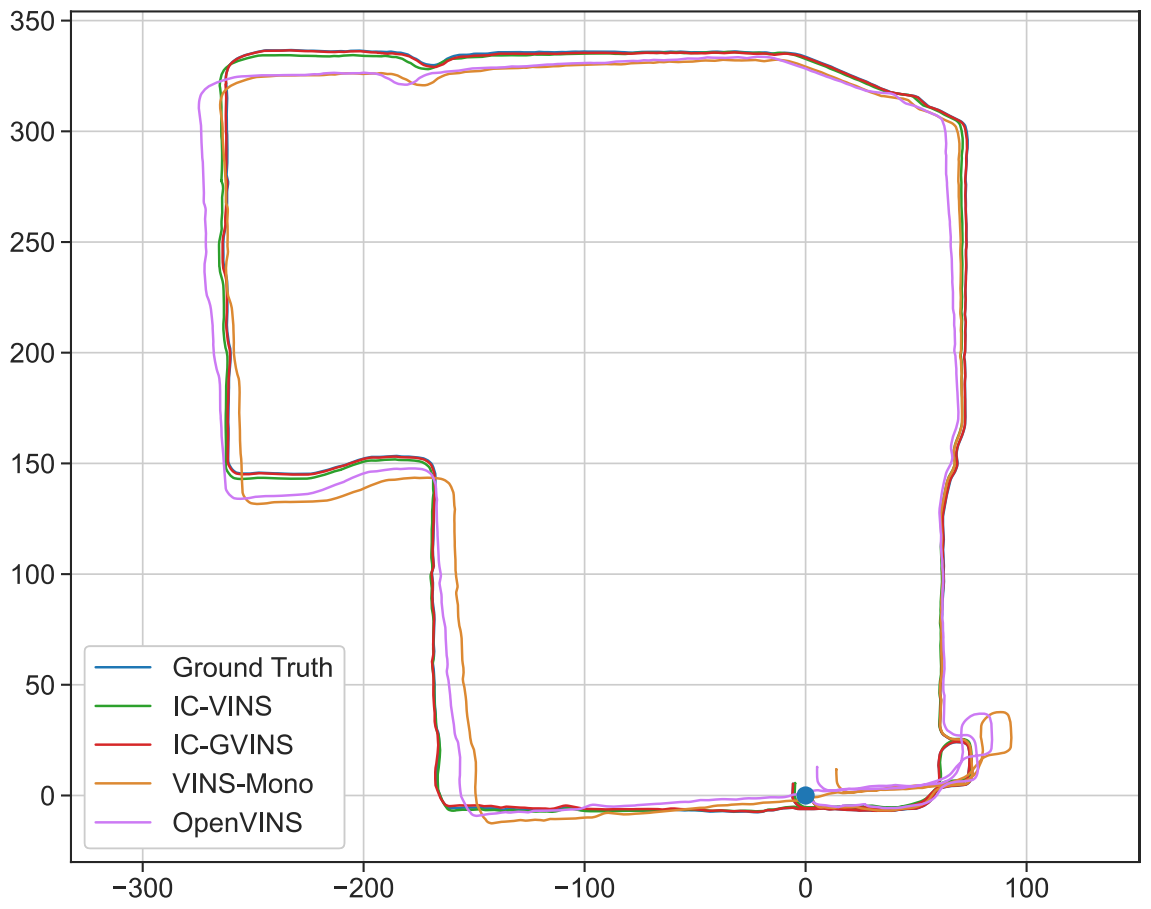

Fig. 9. The trajectories in experiment-3a. The blue circle represents the start point. The trajectory length is 1535 meters.

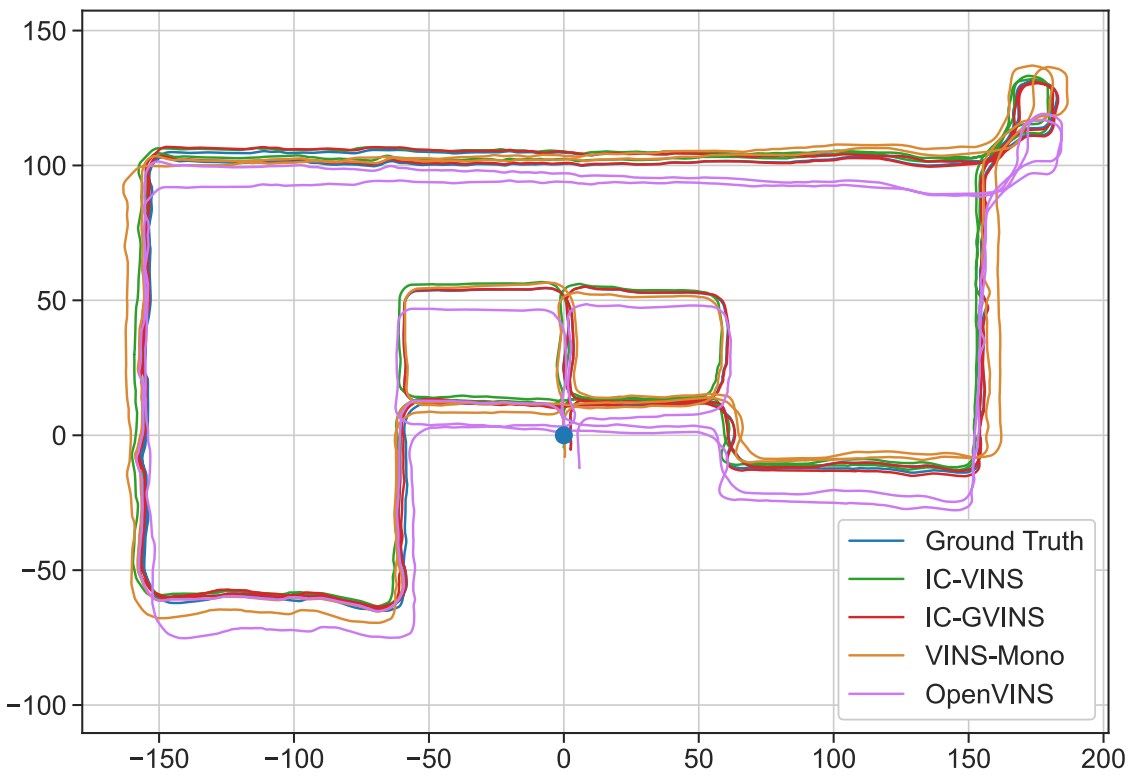

Fig. 10. The trajectories in experiment-3b. The blue circle represents the start point. The trajectory length is 2554 meters.

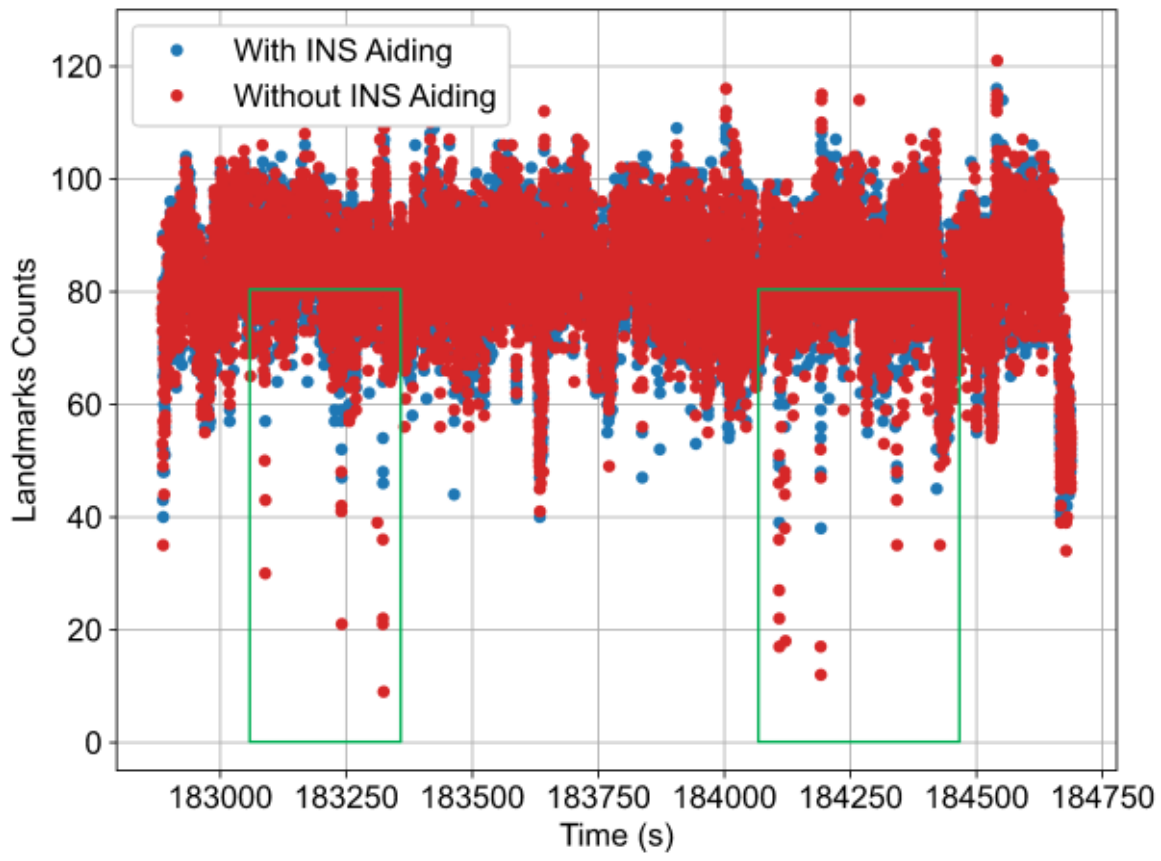

Fig. 11. Comparison of the number of the landmarks in the latest keyframe in experiment-3b. The green rectangles in the figure denote the areas, in where it occurs speed bumps and potholes.

TABLE II
RELATIVE ROTATION AND TRANSLATION ERROR IN EXPERIMENT-3

| RRE / RTE | 50m (deg / %) | 100m (deg / %) | 150m (deg / %) | 200m (deg / %) |
|---|---|---|---|---|
| Experiment-3a | | | | |
| VINS-Mono | 0.38 / 1.72 | 0.58 / 1.29 | 0.78 / 1.21 | 0.97 / 1.27 |
| OpenVINS | 0.55 / 1.39 | 0.79 / 1.20 | 0.96 / 1.26 | 1.12 / 1.34 |
| IC-VINS | **0.19 / 0.53** | **0.24 / 0.44** | **0.29 / 0.39** | **0.34 / 0.38** |
| Experiment-3b | | | | |
| VINS-Mono | 0.33 / 2.17 | 0.46 / 1.87 | 0.54 / 1.71 | 0.60 / 1.61 |
| OpenVINS | 0.55 / 1.49 | 0.89 / 1.42 | 1.18 / 1.46 | 1.47 / 1.55 |
| IC-VINS | **0.20 / 0.78** | **0.29 / 0.63** | **0.35 / 0.56** | **0.40 / 0.52** |

TABLE III
ABSOLUTE ROTATION AND TRANSLATION ERROR

| ARE / ATE | Exp-2 (deg / m) | Exp-3a (deg / m) | Exp-3b (deg / m) |
|---|---|---|---|
| VINS-Mono | 0.724 / 2.052 | 1.847 / 6.877 | 0.655 / 4.610 |
| OpenVINS | 0.527 / 1.241 | 1.247 / 4.470 | 2.338 / 4.228 |
| IC-VINS | 0.468 / 0.877 | 0.532 / 1.205 | 0.350 / 1.642 |
| IC-GVINS | **0.180 / 0.029** | **0.233 / 0.500** | **0.305 / 0.875** |

TABLE IV
AVERAGE RUNNING TIME OF THE IC-GVINS

| PC / On-Board | Exp-1 (ms) | Exp-2 (ms) | Exp-3a (ms) | Exp-3b (ms) |
|---|---|---|---|---|
| Detection and Tracking | 10.5 / 28.0 | 10.4 / 26.9 | 10.5 / 29.2 | 10.6 / 29.3 |
| FGO | 16.8 / 97.1 | 14.2 / 78.6 | 19.5 / 105.8 | 19.0 / 103.1 |

Furthermore, the quantitative results in Table II indicate that the proposed IC-VINS yields superior accuracy, while VINS-Mono and OpenVINS degrade significantly in such challenging environments. The proposed IC-VINS are robust enough to maintain the same accuracy in large-scale challenging environments (experiment-3) as in the less challenging environments (experiment-2). This experiment demonstrates that the proposed INS-centric architecture with the strict outlier-culling strategy is meaningful to improve the robustness and accuracy for VINS in large-scale challenging environments.

## *E. Evaluation of the absolute accuracy*

We also calculated the absolute rotation error (ARE) and absolute translation error (ATE) in experiment-2 and experiment-3 (Exp-2, Exp-3a, and Exp-3b), as showed in Table III. As for the IC-VINS, it outperforms VINS-Mono and OpenVINS in absolute accuracy, because the INS-centric architecture and the precise INS mechanization, which improve the rotation accuracy and further the long-term accuracy. With the help of the GNSS-RTK, the proposed IC-GVINS can achieve centimeter-level positioning in open-sky environment, i.e. experiment-2. The IC-GVINS also exhibits improved accuracy compared to the IC-VINS in challenging environments, i.e. experiment-3, where the accuracy of the GNSS-RTK tends to frequently degrade significantly.

## *F. The effect of the INS aiding in feature tracking*

We compared the number of the landmarks in the latest keyframe in experiment-3b to evaluate the effect of the INS aiding in feature tracking. There are several speed bumps and potholes in the experiment-3b, and they may cause aggressive motion, especially for pitch angle, which make feature tracking extremely challenging. As depicted in Fig. 11, without the INS aiding, the valid landmarks are far fewer than 40 in such cases, and are even close to 0, which means tracking lost. In contrast, the valid landmarks are almost more than 40 during the whole travel with the INS aiding. The results demonstrate that the INS aiding plays an important role in feature tracking, especially in high dynamic scenes, which can improve the robustness of the feature tracking and further improve the robustness of the IC-VINS.

### G. Run time analysis

The average running time of the IC-GVINS in experiment 1~3 (Exp-1~3) are showed in Table IV. All the experiments are running within ROS framework, which demonstrates that the IC-GVINS can perform real-time positioning on both desktop (AMD R7-3700X and 32GB RAM) and on-board computer (NVIDIA Xavier and 32GB RAM).

## V. Conclusions

A robust, real-time, INS-centric GNSS-visual-inertial navigation system for wheeled robot is presented in this letter, which fully utilizes the precise INS information. The prior pose from the INS is incorporated into the keyframe-based visual-inertial pipeline, with strict outlier-culling strategy in both the front-end and back-end. The IMU, visual, and GNSS measurements are tightly fused under the framework of FGO. The GNSS is also employed to perform an accurate and convenient initialization. The proposed method exhibits superior robustness in various visual-degenerated and challenging scenes. Dedicated experiment results demonstrate that the proposed method outperforms the SOTA methods in both robustness and accuracy in various environments.

## Acknowledgement

This research is funded by the National Key Research and Development Program of China (No. 2020YFB0505803), and the National Natural Science Foundation of China (No. 41974024). The authors would like to thank Liqiang Wang, Zaixing Zhang and Guan Wang for the helps in collecting the datasets. The authors would also like to thank Shan Liu for preparing the media materials for this letter.